\documentclass[letterpaper, 10 pt, conference]{ieeeconf}  % Comment this line out if you need a4paper

\IEEEoverridecommandlockouts                              % This command is only needed if 
\usepackage{comment}
\usepackage{graphicx}
\usepackage{balance}
\usepackage{spverbatim}
\usepackage{listings}
\usepackage{subcaption} % for subfigures
\usepackage{float} % for 'H' option
\usepackage{xcolor}
\usepackage{url}
\usepackage{multirow}
\usepackage{fancyvrb}
\usepackage{svg}
\usepackage{comment}

\title{\LARGE \bf Robot Explanation Identity}

\author{Amar Halilovic$^{1}$ and Senka Krivic$^{2}$% <-this % stops a space
\thanks{$^{1}$Amar Halilovic is with the Institute of Artificial Intelligence,
        Ulm University, 89081 Ulm, Germany,
        {\tt\small amar.halilovic@uni-ulm.de}}%
\thanks{$^{2}$Senka Krivic is with the Faculty of Electrical Engineering, University of Sarajevo,
        71000 Sarajevo, BiH,
        {\tt\small senka.krivic@etf.unsa.ba}}%
}

\begin{document}

\maketitle
\thispagestyle{empty}
\pagestyle{empty}

%%%%%%%%%%%%%%%%%%%%%%%%%%%%%%%%%%%%%%%%%%%%%%%%%%%%%%%%%%%%%%%%%%%%%%%%%%%%%%%%%%%%%%%%%%%%%%%%%%%%%%%%%%%%%%%%%%%%%%%%%%%%%%%%%%%%%%%%%%%%%%%%%%%%%%%%%%%%%%%%%%%%%%%%%%%%%%%%%%%%%%%%%%%%%%%%%%%%%%%%%%%%%%%%%%%%%%%%%%%%%%%%%%%%%%%%%%%%%%%%%%%%%%%%%%%%

\begin{abstract}
To bring robots into human everyday life, their capacity for social interaction must increase. One way for robots to acquire social skills is by assigning them the concept of identity. This research focuses on the concept of \textit{Explanation Identity} within the broader context of robots' roles in society, particularly their ability to interact socially and explain decisions. Explanation Identity refers to the combination of characteristics and approaches robots use to justify their actions to humans. Drawing from different technical and social disciplines, we introduce Explanation Identity as a multidisciplinary concept and discuss its importance in Human-Robot Interaction. Our theoretical framework highlights the necessity for robots to adapt their explanations to the user's context, demonstrating empathy and ethical integrity. This research emphasizes the dynamic nature of robot identity and guides the integration of explanation capabilities in social robots, aiming to improve user engagement and acceptance.
\end{abstract}

%%%%%%%%%%%%%%%%%%%%%%%%%%%%%%%%%%%%%%%%%%%%%%%%%%%%%%%%%%%%%%%%%%%%%%%%%%%%%%%%%%%%%%%%%%%%%%%%%%%%%%%%%%%%%%%%%%%%%%%%%%%%%%%%%%%%%%%%%%%%%%%%%%%%%%%%%%%%%%%%%%%%%%%%%%%%%%%%%%%%%%%%%%%%%%%%%%%%%%%%%%%%%%%%%%%%%%%%%%%%%%%%%%%%%%%%%%%%%%%%%%%%%%%%%%%%

\section{Introduction}
Investments in robotics are growing every year \cite{sander2014rise}, and robots could be as ubiquitous in the future as computers are today \cite{breazeal2003toward}. As robots become increasingly integrated into the fabric of society, their roles are expanding beyond traditional industrial applications to encompass more social and interactive functions. This evolution is transforming robots from mere tools to active participants in our daily lives, necessitating a deeper understanding of social dynamics and human-like communication. The emergence of social robots \cite{duffy2000social}, designed to engage with humans in a variety of contexts such as healthcare, education, or customer service, marks a significant shift towards creating machines that can navigate the complexities of human emotions, preferences, and social cues. This shift underscores the importance of developing robots that not only perform tasks but also relate to humans in ways that are intuitive, empathetic, and conducive to building trust and cooperation.

The concept of identity finds diverse applications across a broad spectrum of academic fields, from mathematics and biology to psychology and sociology, as noted by Josselson \cite{josselson1994identity}. Building upon the integration of robots into societal roles, it becomes imperative for social robots to possess a well-defined identity, much akin to human identity theory \cite{erikson1994identity,erikson1994identity1} that emphasizes the multifaceted nature of self-concept and social representation. Just as personal and social identities shape human interactions, the identity of a robot could significantly influence how it is perceived and engaged with humans. Moreover, besides the social, robots have the potential to possess personal identities as well \cite{alonso2023can}. Drawing parallels with the fluidity present in human identities \cite{smedley1998race}, which adapt and evolve in response to different contexts and social settings, robot identities must also exhibit a similar level of adaptability. This fluidity allows robots to seamlessly transition between roles and environments, mirroring the dynamic nature of human social interactions. Developing such adaptable robot identities necessitates a nuanced understanding of the core principles of human identity, enabling robots to better align with human expectations, foster empathetic connections, and ultimately enhance the synergy between humans and themselves in shared spaces.

With the increasing employment of machine learning in robotics, the underlying algorithms driving robot behavior grow increasingly complex, often rendering their actions opaque to the humans they are designed to interact with. This complexity emphasizes the pressing need for effective explanation mechanisms within robotics, ensuring that users can understand the rationale behind a robot's decisions and actions. Accessible explanations are essential not only for transparency but also for building user trust and facilitating smoother human-robot collaborations. Explanations could help bridge the gap between sophisticated robotic functionalities and user comprehension, thereby enhancing the overall efficacy and acceptance of robots in various social and professional settings. In response to the challenge of making robot behaviors understandable, we introduce the concept of Explanation Identity as an interdisciplinary framework for when robots need to articulate their actions and decisions. This identity is not merely a technical proposal; it embodies and integrates both the technical aspects of explanation generation and the social nuances of how they are communicated to users. Thus it is multidisciplinary.

%%%%%%%%%%%%%%%%%%%%%%%%%%%%%%%%%%%%%%%%%%%%%%%%%%%%%%%%%%%%%%%%%%%%%%%%%%%%%%%%%%%%%%%%%%%%%%%%%%%%%%%%%%%%%%%%%%%%%%%%%%%%%%%%%%%%%%%%%%%%%%%%%%%%%%%%%%%%%%%%%%%%

\section{Identity in Robotics}
The concept of identity, especially artificial identity, requires careful consideration, distinct from personal or animal identities \cite{digiovanna201720,babushkina2018artificial}. Identity formation is a critical link between the individual and society \cite{erikson1994identity}, a process that is inherently transdisciplinary \cite{hammack2008narrative} and multifaceted, encompassing personal, social, and place identities \cite{twigger2003identity}. Moreover, identity is dynamic and can be either assigned or chosen \cite{adams1996developmental}, influencing and being influenced by surrounding living systems. Alonso \cite{alonso2023can} posits that robots could have personal identities with unique characteristics.
The exploration of identity in robotics, particularly in the realm of social robots, presents significant implications for the design and interaction paradigms of such systems. The study of robot identity is still an emerging area \cite{seabornidentified}, yet critical for enhancing robot effectiveness in social settings \cite{winkle2021flexibility}. The design of personalized robot identities introduces the possibility of robots possessing adaptable social cues, enabling them to modify aspects of their identity in real time to align with varying contexts or interaction goals. This adaptability, unparalleled in humans, prompts considerations about user perceptions of such changes and the extent to which a robot can alter its identity before being perceived as an entirely new entity \cite{winkle2021flexibility}.
Furthermore, the concept of agent migration, where a robot's identity can be transferred between different physical embodiments or across devices, extends the versatility of personalized robot identities \cite{imai1999agent,misikangas2000agent,ho2009initial}. This raises discussions on the continuity of identity and the coherent projection of a robot's social persona, irrespective of its physical form, as explored in \cite{koay2009user,kriegel2010digital,reig2019leveraging,reig2020not}. The study by Straub et al. \cite{straub2010incorporated}, on the other hand, illustrates how users can attribute distinct identities to robots that are independent of the identity of the robot's human operator, underlining the social presence influence the robots can convey to humans around them.

In the broader context of Human-Robot Interaction (HRI), Nass et al. \cite{nass1997machines} establish that humans inherently perceive robots as social beings, suggesting that interactions with robots can influence cultural and social norms similar to Human-Human Interactions (HHI). This is aligned with the view of Miranda et al. \cite{miranda2023examining} who assert that social identity is shaped not within an individual but through the perceptions of others, challenging the conventional 1-1-1 mapping of mind, body, and identity typically associated with humans \cite{jackson2021design}. This paradigm shift invites a reevaluation of identity constructs in robotics, emphasizing the need for designs that accommodate the fluidity and multiplicity of identity in social interactions.

%%%%%%%%%%%%%%%%%%%%%%%%%%%%%%%%%%%%%%%%%%%%%%%%%%%%%%%%%%%%%%%%%%%%%%%%%%%%%%%%%%%%%%%%%%%%%%%%%%%%%%%%%%%%%%%%%%%%%%%%%%%%%%%%%%%%%%%%%%%%%%%%%%%%%%%%%%%%%%%%%%%%

\section{Explanations and Humans}
To understand the role and significance of explanations in robotics, it is essential to first understand the dynamics of how humans define and use explanations.
Human explanation theory encompasses a variety of perspectives, with the common notion that explanation is a multifaceted concept that is reflective of diverse cognitive styles and individual differences among people. Kahneman \cite{daniel2017thinking} highlights that humans engage in distinct modes of thinking, suggesting that the nature and depth of explanations can influence the mode of reasoning employed. Moreover, explaining how the mind works through symbolic and informational terms is generally distinct from explaining how the brain works on a physical, nerve-based level \cite{simon1992explanation}. Nevertheless, there tends to be a consensus regarding the types of explanations favored by humans; namely, those that are highly likely and simple \cite{lombrozo2007simplicity}. When explaining, individuals typically resort to causal structures \cite{lombrozo2006structure} and often consider counterfactual, or "what-if," scenarios \cite{keil2006explanation}.
Pragmatists define explanation as an answer to why questions, as outlined by van Fraassen \cite{van1988pragmatic}. Malle, in his body of work \cite{malle2001folk,malle2005folk,malle2006mind}, ties the concept of explanation to the folk theory of intentionality, distinguishing between intentional and unintentional behaviors. He posits that explanations can be framed in terms of reasons for intentional actions and causes for unintentional ones. Furthermore, Graaf and Malle argue in \cite{de2017people} that autonomous systems ought to provide explanations for their actions following the same theory. While the theory of intentionality is widely recognized within explanation theories, it faces notable criticism, such as Lovett's argument \cite{lovett2006rational}. Lovett contends that a causal explanation model is superior to the intentional one, while also emphasizing the significance of understanding explanations from a functional perspective.

%%%%%%%%%%%%%%%%%%%%%%%%%%%%%%%%%%%%%%%%%%%%%%%%%%%%%%%%%%%%%%%%%%%%%%%%%%%%%%%%%%%%%%%%%%%%%%%%%%%%%%%%%%%%%%%%%%%%%%%%%%%%%%%%%%%%%%%%%%%%%%%%%%%%%%%%%%%%%%%%%%%%

\section{Explanations in Robotics}
%The exposure to autonomous robots led to more negative attitudes towards robots broadly and greater resistance to robotics research, in contrast to reactions to non-autonomous robots \cite{zlotowski2017can}.
Even minor deceptive tactics in robot movements, such as faking an action towards one object before selecting another, can diminish human trust \cite{dragan2014analysis}. The increasing demand for robots to be transparent, trustworthy, and explainable is well recognized \cite{de2018explainable, anjomshoae2019explainable}, leading to significant efforts towards creating transparent robots \cite{winfield2021ieee}. A crucial aspect of making robots understandable is how they communicate their intentions; failures in this communication can make robots appear unsettling to people, regardless of their decision-making accuracy \cite{lomas2012explaining, williams2015covert}. Studies indicate that providing explanations enhances both user comprehension \cite{kwon2018expressing,wang2018my} and trust in the system \cite{stange2020effects,das2021explainable,ambsdorf2022explain,lyons2023explanations}. Explanations need to be flexible \cite{ribera2019can} and to include an adequate amount of detail \cite{han2021need}.

Current strategies in explainable robotics align with the broader Explainable AI (XAI) framework \cite{xu2019explainable}, utilizing model-specific and model-agnostic methods. Specialized domains are Explainable AI Planning (XAIP) \cite{fox2017explainable} and Explainable Reinforcement Learning (XRL) \cite{puiutta2020explainable}, where the focus is on explaining different robot applications and games \cite{sieusahai2021explaining, he2021explainable, remman2021robotic}. %\cite{sakai2022explainable, sieusahai2021explaining, he2021explainable, remman2021robotic}. 
There is a lack of research on what kind of identity robots should have when explaining their actions to humans. The only piece of work conceptually close to the identity in explainable robotics is by Chakraborti et al. \cite{chakraborti2017plan, chakraborti2019plan}. They introduce the concept of Model Reconciliation in Explainable Planning and describe it as aligning the mental model of a system's behavior held by a user with the actual operational model of the system. In simpler terms, it is about ensuring that the user's understanding of how and why a robot makes certain decisions or plans aligns closely with the underlying mechanisms of that robot. They assume that the robot's mental model is given and exists. We propose an agnostic model of building blocks and features of robot mental models resembling a robot's explanation identity.

%%%%%%%%%%%%%%%%%%%%%%%%%%%%%%%%%%%%%%%%%%%%%%%%%%%%%%%%%%%%%%%%%%%%%%%%%%%%%%%%%%%%%%%%%%%%%%%%%%%%%%%%%%%%%%%%%%%%%%%%%%%%%%%%%%%%%%%%%%%%%%%%%%%%%%%%%%%%%%%%%%%%

\section{Design of Robot Explanation Identity}

Collaboration between humans and robots is pivotal for maximizing performance in task execution \cite{wilson2018collaborative}, with interpersonal communication and relationship building being the main factors for success in such collaborations \cite{urakami2021building}. However, the cognitive capacity of humans suggests a limit to how much information can be held in working memory at once \cite{miller1956magical}, indicating that explanations provided by robots should be mindful of these human limitations. Coeckelbergh \cite{coeckelbergh2010moral} cautions against placing excessive demands on robots beyond what is expected of humans, emphasizing the need for realistic expectations. Furthermore, Babushkina and Votsis \cite{babushkina2018artificial} advocate for integrating users' stakes with psychological and ethical considerations in designing 'smart' technologies for HRI.

To design a robot explanation identity, one must account for the traits an explaining robot (explainer) should have for successful explanation generation and delivery. In our proposed framework for explainable robot navigation \cite{halilovic2023influence}, we model the explanation generation as a hierarchical process with four key steps defining \textit{what?}, \textit{when?}, \textit{how?} and \textit{how long?} to explain. Our framework is robot and explanation-agnostic. We propose elevating these four steps to the global robot explanation generation level, which is agnostic to the specific robot behavior or task:

\begin{itemize}
\item \textit{What to explain } corresponds to the explanation target problem. Irrespective of whether the robot navigates or manipulates objects, it should be able to choose the explanation object in a given setting. Usually, it is linked to the last mistake it made, whether it is a planning failure or deviation from an expected behavior.

\item \textit{When to explain} is a problem of explanation timing, i.e., choosing the suitable moment and context for communicating an explanation. Humans, knowledgeable of the social norms of other people, usually judge when to communicate something. Misjudgment can lead to unwanted reactions from explanation recipients (explainees) that can lead to a decrease instead of an increase in trust in explainers.

\item  \textit{How to explain} decides on the modality of explanation. How to represent explanations differs on contextual and environmental factors and explainee conditions. Employed modalities in explainable robotics include visual, textual, and verbal modalities, as well as expressive motions and lights. Combining two or more modalities gives multimodal, i.e., hybrid approaches, which allow for contextually richer explanations, which may be desirable in certain situations.

\item \textit{How long to explain} addresses both the duration and the length of the explanation, which are heavily interaction-dependent. The explanation can be inherently long, but the explainee may want a shorter version. On the other hand, the explanation itself can be short, but the user may wish for a more detailed explanation or several repetitions of it (due to misunderstanding), which can make the whole explanation process durate longer.

\end{itemize}
\begin{figure}[ht]
\centering
\includegraphics[width=0.47\textwidth]{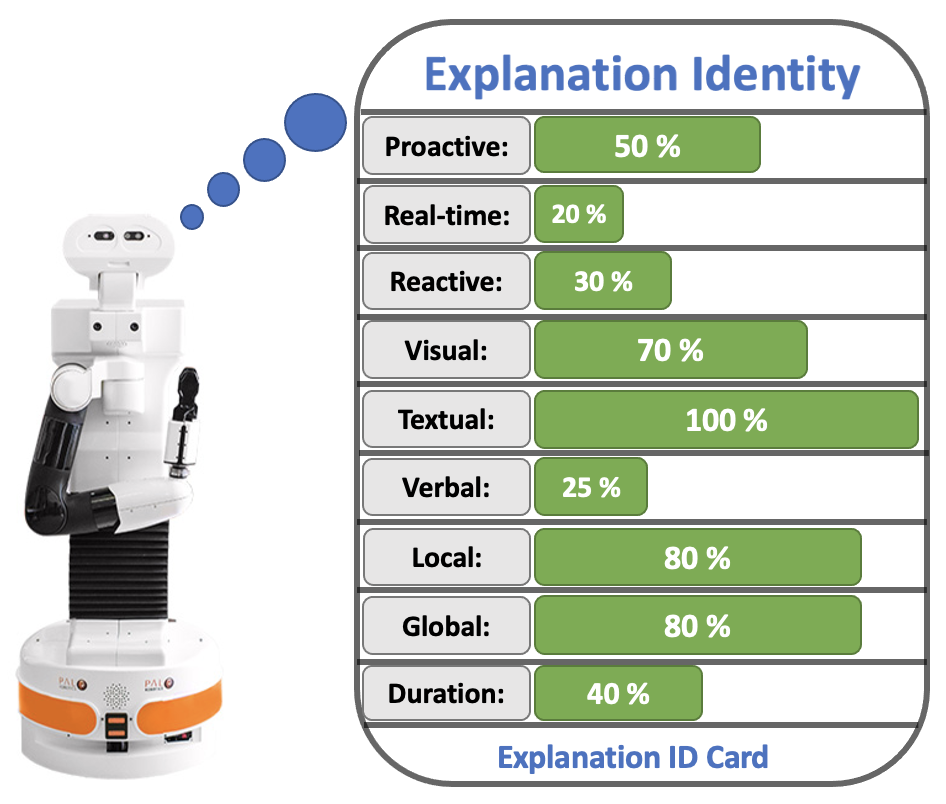}
\caption{
The robot (a TIAGo from PAL robotics \cite{pages2016tiago}) has an explanation identity as a part of its personality. Its explanation identity forms an explanation ID card, defined by different explanation-important variables (features), which are estimates of a user's (explainee) preferences at an explanation time. Each variable is accompanied by the percentage (probability, value), which defines the probability of whether that explanation feature will be activated for a specific user at a specific time. The probabilities of features fluctuate over time and are dependent on the environment, context, explainees, and the robot's previous explanation identity state. One approach for adapting these probabilities is through life-long (open-ended) learning, where through interactions the robot continuously adapts its initially learned (usually with supervised learning) explanation identity based on data from sensor inputs and its current explanation identity state.
} 
\label{fig:explanations}
\end{figure}

We foresee these four explanation phases as the cornerstone elements that constitute the essence of the robotic explanatory identity. These phases are intrinsically sequential but also possess a hierarchical nature, which enhances the structural complexity and diversity of explanatory approaches. Each phase is characterized by a distinct set of potential attributes, as illustrated in Figure \ref{fig:explanations}. The explanation target (the aspect to be explained) is typically predetermined by the contextual backdrop of the robot's activities at the start of the explanation generation process. Concurrently, the scope of the explanation outlines the extent to which the explanation will be either \textit{local} or \textit{global}. Nonetheless, these categories can be mutually exclusive, as an explanation recipient may require explanations that include both local and global perspectives simultaneously, thereby obliging the robot to include explanations that span both scopes. Should the robot deduce that the explanation recipient's requirement is confined to a local explanation, devoid of a global perspective, it will accordingly decrease the likelihood of providing a global explanation. Moreover, the modalities of explanation representation prescribe the medium through which the explanation is communicated to the recipient. Typically, a robot distinguishes among \textit{visual}, \textit{textual}, and \textit{verbal} modes of explanation. An unimodal representation is exclusively visual, textual, or verbal, whereas a multimodal representation encompasses a combination thereof. Given that a recipient may prefer a blend of these modalities, these options must remain disjunct (see Figure \ref{fig:explanations}). The temporal aspect of an explanation generally pertains to the timing of the explanation relative to an action, categorizable as \textit{proactive} (before an action), \textit{real-time} (concurrent with an action), or \textit{reactive} (after an action). In certain scenarios, it may be necessary for the robot to provide explanations both before and after an action. Additionally, the \textit{duration} of an explanation—which encompasses both its actual size and the time required to convey it—ought to be modulated by the preferences of the recipient.

Robot explanation identity should also be fluid so the robot can enhance and restructure its explanations, focusing only on the steps it needs to change without the need to go through the whole explanation generation again. These questions are also essential to the problem of human communication, making its explanation of identity strongly tied to the communicational mental model. Still, robot explanation identity should serve as a distinct identity, as robot communication capabilities are still not close to the complexity of those of humans. It has been noted that robots face criticism even when they do not make errors \cite{kim2006should}. To reduce unjustified criticisms, robot explanations should be personalized \cite{anjomshoae2019explainable}. For example, an explanation provided to a software engineer might differ significantly from one provided to someone with no technical background. The robot should tailor its explanations based on the interaction context and the user's background knowledge.

%%%%%%%%%%%%%%%%%%%%%%%%%%%%%%%%%%%%%%%%%%%%%%%%%%%%%%%%%%%%%%%%%%%%%%%%%%%%%%%%%%%%%%%%%%%%%%%%%%%%%%%%%%%%%%%%%%%%%%%%%%%%%%%%%%%%%%%%%%%%%%%%%%%%%%%%%%%%%%%%%%%%

\section*{Conclusions}
We have highlighted the dynamics of human-robot collaboration, emphasizing the necessity of human-robot communication and relationship-building to achieve optimal task performance. The complexity of robot behavior and the cognitive limitations of humans necessitate the design of human-interpretable explanations from robots aware of these limitations. The exploration of identity in explainable robotics reveals that robots may need unique identity characteristics, distinct from human and animal attributes, to explain efficiently. The design of HRI systems should adopt these implications. By integrating social, psychological, and ethical considerations, alongside acknowledging the dynamic and transdisciplinary nature of identity, there is a potential to create a robot explanation identity that is distinct but still intertwined with other parts of the robot persona. For example, imagine a robot that needs to pass between two standing people talking to each other. A robot without an explanation identity may do it abruptly even though it does not have secret malicious intentions. It would not be able to explain its intention before, during, or after the passing, although the people impacted by its action may require an explanation. Even if such a robot respects human social norms, it would not be able to proactively explain its intention and would stay frozen in place. However, the robot with an explanation identity, regardless of whether it respects or does not respect human social norms, would be able to explain its action, either proactively, reactively, or during the action, which may decrease the negative impact on the people involved in the interaction with the robot. We outline the first conceptual design of an explanation identity, which is parametrized and allows for hierarchical and sequential explanation generation. The proposed concept of robot explanation identity is multi-faceted, encompassing technical challenges, philosophical questions, and societal implications. The future work will explore human explanation preferences and subsequential preferences-driven redesign of the proposed robot explanation identity design and its embodiment on a real robot. Producing human-friendly explanations paves the way for more effective, empathetic, and ethically informed collaborations between humans and robots. As AI and robotics continue to evolve, so will how robots explain their decisions, profoundly impacting HRI and the role of robots in society.  

%%%%%%%%%%%%%%%%%%%%%%%%%%%%%%%%%%%%%%%%%%%%%%%%%%%%%%%%%%%%%%%%%%%%%%%%%%%%%%%%%%%%%%%%%%%%%%%%%%%%%%%%%%%%%

%\section*{Author Biographies}
%\paragraph{\textbf{Amar Halilovic} is a PhD Student and research and teaching assistant at Ulm University, Germany. His research is on Explainable Robotics. He is focusing on developing methods that make the decision-making processes in robot motion planning interpretable.}

%\paragraph{\textbf{Senka Krivic} is an Assistant Professor at the Faculty of Electrical Engineering at the University of Sarajevo. Her research interests are in Robot Learning, AI Planning and Reasoning, Trusted and Explainable Autonomous Systems, Human-Robot Interaction, Machine Learning, and Optimization.} 

%%%%%%%%%%%%%%%%%%%%%%%%%%%%%%%%%%%%%%%%%%%%%%%%%%%%%%%%%%%%%%%%%%%%%%%%%%%%%%%%%%%%%%%%%%%%%%%%%%%%%%%%%%%%%%%%%%%%%%%%%%%%%%

\balance
\bibliographystyle{IEEEtran}
\bibliography{./sources}

%%%%%%%%%%%%%%%%%%%%%%%%%%%%%%%%%%%%%%%%%%%%%%%%%%%%%%%%%%%%%%%%%%%%%%%%%%%%%%%%%%%%%%%%%%%%%%%%%%%%%%%%%%%%%%%%%%%%%%%%%%%%%%

\end{document}